\documentclass[runningheads]{llncs}

\newif\ifready
\readytrue
\usepackage[mobile]{eccv}

\usepackage{eccvabbrv}

\usepackage{graphicx}
\usepackage{booktabs}

\usepackage[accsupp]{axessibility}  
\ifready
\else
    \usepackage[pagebackref,breaklinks,colorlinks]{hyperref}
\fi
\usepackage{orcidlink}

\usepackage{colortbl}
\usepackage{multirow}
\usepackage{hhline}

\usepackage{amsmath}
\DeclareMathOperator*{\argmax}{arg\,max}

\usepackage{wrapfig}
\usepackage{soul}
\usepackage{ulem}
\definecolor{Cerulean}{rgb}{0,0,0.95}
\definecolor{LimeGreen}{rgb}{0.15,0.65,0.15}
\definecolor{RoyalBlue}{rgb}{0.25,0.41,0.88}
\definecolor{Rose}{rgb}{1.0, 0.15, 0.21}
\definecolor{Orange}{rgb}{1.0, 0.5, 0.0}
\definecolor{Gray}{gray}{0.6}
\definecolor{Black}{gray}{0.0}
\definecolor{Purple}{rgb}{0.77,0.12,0.64}

\begin{document}

\title{SeFlow: A Self-Supervised Scene Flow \\
Method in Autonomous Driving}


\author{Qingwen Zhang\inst{1}\orcidlink{0000-0002-7882-948X} \and
Yi Yang\inst{1,2}\orcidlink{0000-0002-6679-4021}  \and
Peizheng Li\inst{3,4}\orcidlink{0000-0003-2140-4357} \and \\
Olov Andersson\inst{1}\orcidlink{0000-0001-7248-1112} \and
Patric Jensfelt\inst{1}\orcidlink{0000-0002-1170-7162}}

\authorrunning{Qingwen~Zhang, Yi~Yang, Peizheng~Li et al.}

\institute{RPL, KTH Royal Institute of Technology, Stockholm, Sweden \and
Scania CV AB, Södertälje, Sweden \and
Mercedes-Benz AG, Sindelfingen, Germany \and
University of Tübingen, Tübingen, Germany
}

\maketitle
\begin{abstract}
Scene flow estimation predicts the 3D motion at each point in successive LiDAR scans. This detailed, point-level, information can help autonomous vehicles to accurately predict and understand dynamic changes in their surroundings.
Current state-of-the-art methods require annotated data to train scene flow networks and the expense of labeling inherently limits their scalability. 
Self-supervised approaches can overcome the above limitations, yet face two principal challenges that hinder optimal performance: point distribution imbalance and disregard for object-level motion constraints. 
In this paper, we propose SeFlow, a self-supervised method that integrates efficient dynamic classification into a learning-based scene flow pipeline. We demonstrate that classifying static and dynamic points helps design targeted objective functions for different motion patterns. We also emphasize the importance of internal cluster consistency and correct object point association to refine the scene flow estimation, in particular on object details. 
Our real-time capable method achieves state-of-the-art performance on the self-supervised scene flow task on Argoverse 2 and Waymo datasets. 
The code is open-sourced at \url{https://github.com/KTH-RPL/SeFlow}.

\end{abstract}

\section{Introduction}
Scene flow~\cite{vedula2005three} captures the 3D velocity at every point in a point cloud. These detailed 3D flow estimates can enhance downstream tasks in autonomous driving, such as detection, segmentation, tracking, and occupancy flow estimation~\cite{najibi2022motion}. 
A common paradigm for addressing the scene flow problem is supervised learning by utilizing annotated LiDAR data~\cite{fastflow3d, zhang2024deflow}.
However, expensive labeling inherently limits the scalability of supervised learning methods.

\begin{figure}[t!]
\centering
\includegraphics[trim=100 170 150 100, clip, width=\linewidth]{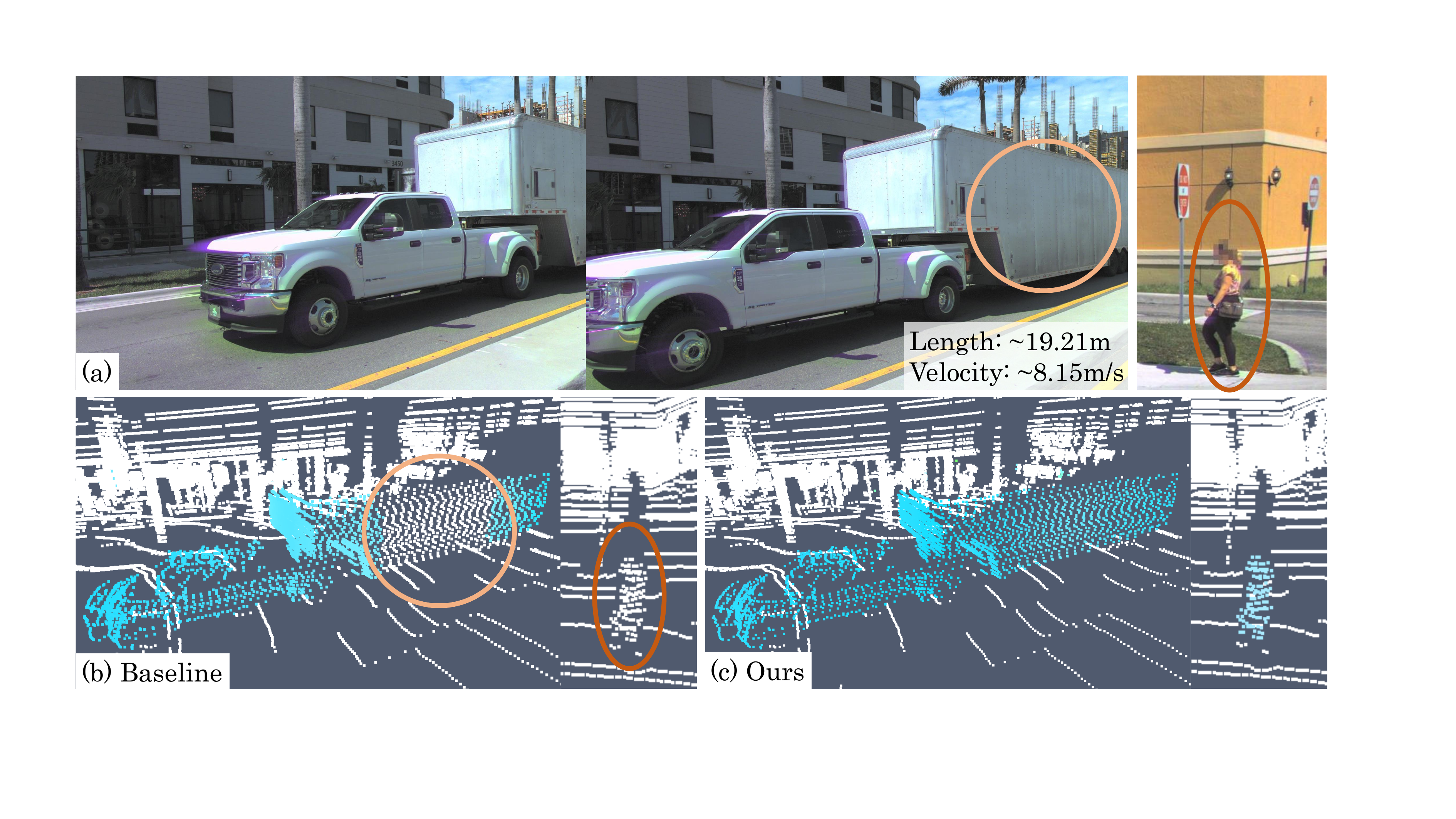}
\caption{LiDAR scene flow estimation using our SeFlow method on Argoverse 2. 
The predicted scene flow for each point is color-coded based on direction. The white indicates static points whose flow is zero. More saturated colors indicate higher velocities. 
(a) Camera view for visualization purposes only. (b),(c) are zoomed-in views showing the baseline from ZeroFlow~\cite{zeroflow} as well as SeFlow (ours). 
When predicting the flow of a large and long vehicle, the baseline predicts a portion of the flow as 0, whereas our estimates are consistent. In addition, the baseline tends to ignore small-scale objects, e.g., pedestrians, while our method can better handle such small and slow-moving objects.
}
\label{fig:cover}
\vspace{-1.0em} 
\end{figure}

Given the difficulty and expense of labeling scene flow ground truth, we instead focus on self-supervised scene flow approaches.
Existing self-supervised methods can be divided into two categories: knowledge distillation and data exploration. 
Knowledge distillation methods~\cite{aleotti2020learning,zeroflow,song2023knowledge} typically rely on a `teacher' method to generate pseudo flow labels. These pseudo labels are then used to supervise student models.
Data exploration methods~\cite{mittal2020just,baur2021slim,li2023fast,li2021neural,vidanapathirana2024multi}, on the other hand, directly utilize the predicted flow to project the input points and find nearest neighbors in the next frame to establish constraints for training.

A key challenge for self-supervised methods is that the vast majority of the points are static (about 86\% of points are background~\cite{Argoverse2_2021, chodosh2023re}). 
This data imbalance often leads to overly conservative scene flow predictions.
One common way to counter this is to use large amounts of training data as in ZeroFlow~\cite{zeroflow}. 
ZeroFlow experimentally shows that more data helps to improve the dynamic flow accuracy.
In this paper, we investigate ways to improve data efficiency and take inspiration from supervised methods~\cite{fastflow3d,zhang2024deflow}. We address the data imbalance by first classifying points as dynamic or static based on traditional ray casting when integrating frames~\cite{daniel2024dufomap}. 
This allows us to constrain the corresponding motion patterns by proposing novel loss functions.

Another shortcoming of existing self-supervised scene flow methods is that they disregard object-level motion constraints. 
The flow should be consistent, i.e., all points in a rigid object should have similar flows. 
A clear case of inconsistent flow can be seen on the big truck in \cref{fig:cover}, where ZeroFlow~\cite{zeroflow} predicts an absence of flow in certain sections. 
This inconsistency is caused by incorrect object point associations which also can be observed in small-scale objects, as shown in the case of the pedestrian. 
To account for object-level motion constraints, we propose to cluster the dynamic points into object candidates and encourage consistent flow and correct associations for the points in each cluster. This reduces the fragmentation of the flow estimate for points inside the same object. 

Overall, our method improves the estimated flow accuracy and addresses issues in previous methods as shown in \cref{fig:cover}(c). We propose an efficient and effective self-supervised scene flow method that integrates traditional classification and learning-based strategies.
\ifready
    Our approach is open-source at \color{blue}\url{https://github.com/KTH-RPL/SeFlow}\color{black}. 
\fi
Our primary contributions include:
\begin{itemize}
\item We propose SeFlow, a novel method that integrates a dynamic classification method in formulating efficient self-supervision objectives.
\item We further construct loss functions to learn dynamic flow estimation in imbalanced data and ensure consistent object-level flow, mitigating the effects of correspondence errors.
\item We show that SeFlow achieves state-of-the-art results on the self-supervised scene flow task in Argoverse 2 and Waymo datasets and even outperforms all but one supervised method on the leaderboard.
\end{itemize}

\section{Related Work} 
In this section, we explore existing works in scene flow estimation. We also discuss current traditional frameworks capable of classifying dynamic points, highlighting their relevance and application in the context of scene flow tasks.
\label{sec:related_work}
\subsection{Scene Flow Estimation}
Scene flow estimation in autonomous driving is slightly different from flow estimation in object registration.
Methods in object registration~\cite{wei2020pv,wang2023dpvraft,shen2023self,scoop,liu2023difflow3d,deng2023rsf} focus on relatively small-scale point cloud data like synthetic datasets Shapenet~\cite{chang2015shapenet} and FlyingThing3D~\cite{mayer2016large}. 
These methods scale poorly with the number of points. When applied to point cloud data for autonomous driving, they require downsampling to 8,192 points or less~\cite{fastflow3d,menze2015object}.
In recent datasets like Argoverse 2~\cite{Argoverse2_2021} and Waymo~\cite{fastflow3d}, the number of points in one full frame is around 80k-177k.
Methods that can handle such large-scale point cloud data as input usually use a voxel-based pipeline~\cite{fastflow3d,zhang2024deflow}. 
However, expensive labeling of ground truth flow limits the scalability of these supervised methods, especially in autonomous driving where we have continuously increasing amounts of potential training data.

To train models without labeled data, recent methods propose self-supervised losses.
Many commonly used losses, such as Chamfer distance, are based on the nearest neighbor distance between two point cloud inputs~\cite{kittenplon2021flowstep3d,tishchenko2020self,wu2020pointpwc,mittal2020just,baur2021slim,li2021neural,li2023fast}.
However, a major limitation of nearest neighbor based losses is that only part of the points on dynamic objects provides good supervision due to incorrect associations. This is especially apparent when big trucks are moving fast in autonomous driving scenarios (see \cref{fig:cover} and \cref{fig:overlap_area}).
Mittal \textit{et al.} \cite{mittal2020just} define a self-supervised loss by tracking a patch forward and backward in time to form a cycle while penalizing the errors through cycle consistency and feature similarity. 
Baur \textit{et al.} \cite{baur2021slim} propose three loss functions with k-NN loss, rigid cycle consistency inspired by Mittal, and artificial labels based on the distance between points in the original and estimated point clouds. 
In the following we describe NSFP~\cite{li2021neural}, FastNSF~\cite{li2023fast}, and ZeroFlow~\cite{zeroflow} in more detail. These are the publically available methods that perform best on the Argoverse 2 self-supervised scene flow challenge and are therefore used as our baselines.

NSFP~\cite{li2021neural} is an optimization-based method. For each pair of consecutive input point clouds, NSFP iteratively learns new weights for an MLP network to predict the flow using the Chamfer distance loss. 
However, thousands of iterations are needed and their runtimes extend from 26 to 35 seconds per frame, which fails to meet the real-time requirements of autonomous driving.
FastNSF \cite{li2023fast} improves the efficiency by voxelizing the point cloud first and then converting it to a distance transform map for faster neighbor calculation. 
This reduces the runtime to 0.5 seconds, at the expense of increased error.

ZeroFlow \cite{zeroflow} adopts a semi-supervised strategy. 
Pseudo-flow labels are created offline using NSFP~\cite{li2021neural}, and a FastFlow3D~\cite{fastflow3d} model is used as a student for knowledge distillation. This setup allows the student model to perform real-time inference.
In this case, the teacher network needs significant resources (around 3.6 GPU months~\cite{zeroflow}) to label the entire training data. The final accuracy is also influenced by the performance of the teacher.

To increase data efficiency and address the limitations of the above methods, we propose to integrate efficient ray-casting-based dynamic awareness mapping into our pipeline. 
The core idea is to classify which points move and cluster these points into objects for which we can estimate group-level motion statistics and define better self-supervised loss functions.

\subsection{Dynamic Awareness in Mapping}
In the field of Simultaneous Localization and Mapping (SLAM), there is a growing interest in dynamic awareness~\cite{zhang2023benchmark,schmid2023dynablox,daniel2024dufomap,wu2024moving,pfreundschuh2021dynamic}. This interest stems from the fact that points associated with moving objects in a scene can significantly impact localization and planning performance. 
Existing dynamic awareness methods in mapping \cite{daniel2024dufomap,schmid2023dynablox,zhang2023benchmark} often utilize ray casting techniques to binary classify points as either dynamic or static.

The scene flow task, on the other hand, aims to predict the specific 3D flow at each point, a goal that extends beyond the mere categorization of dynamic and static points.
In mapping, a point is considered dynamic if it moves once within a scene (even if it becomes static later), whereas, in scene flow tasks, a point is deemed dynamic if it moves faster than a certain velocity threshold in the current frame. 
Despite these differences, insights from the mapping field are invaluable. By integrating information over time, these frameworks develop a more comprehensive understanding of the entire scene. Such scene level dynamic awareness can inform and enhance our exploration of point cloud data in scene flow tasks.

\section{Problem Statement}
This work addresses the problem of real-time scene flow estimation in autonomous driving. Given two consecutive input point clouds, $\mathcal{P}_t$ and $\mathcal{P}_{t+1}$, along with the ego movement $\mathbf{T}_{t,t+1} \in SE(3)$, the goal is to predict the motion vector as flow $\hat{\mathcal{F}}_{t,t+1}(p) = (x,y,z)^T$ for each point $p \in \mathcal{P}_t$.

The objective is to minimize the End Point Error (EPE) which represents the difference between the predicted flow and the ground truth flow, as expressed by the following equation:
\begin{equation}
    \min \underbrace{ \frac{1}{|\mathcal{P}_t|} \sum_{p \in \mathcal{P}_t} \left\| \hat{\mathcal{F}}(p) - \mathcal{F}_{gt}(p) \right\|_2 }_{\text{EPE}},
    \label{eq:opti}
\end{equation}
where $|\mathcal P_{t}|$ denotes the cardinality of (i.e., the number of points in) $\mathcal P_t$. For consistency, in the subsequent presentation, capital symbols correspond to sets, while the lowercase symbols represent variables of specific points.

\section{Method}
\begin{figure}[h]
\centering
\includegraphics[trim=250 150 250 150, clip, width=\linewidth]{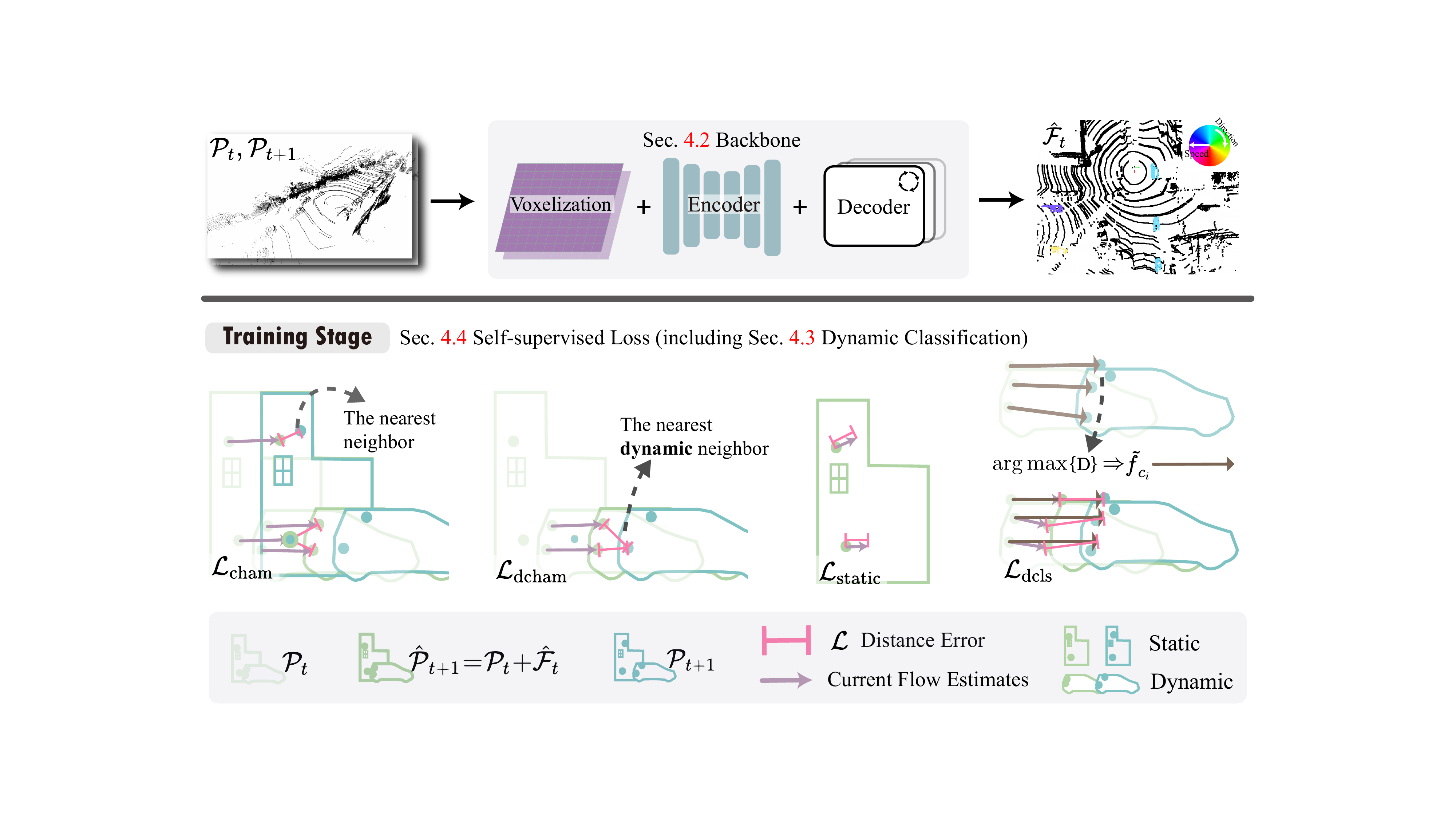}
\caption{
\textbf{SeFlow Architecture}. 
Top: With two consecutive point clouds as inputs, our model predicts the estimated flows of all points. 
Bottom: Conceptual visualization of the Chamfer loss and the three proposed training losses. With the original input $\mathcal P_t$ (2 static points for the building, and 2 dynamic points for the car) plus the estimated flow $\hat {\mathcal F}_t$, we can calculate the error between estimated $\hat{\mathcal P}_{t+1}$ and the next frame point cloud $\mathcal P_{t+1}$ ($\mathcal L_{\text{cham}}$). 
The second part is $\mathcal L_{\text{dcham}}$ that only calculates the distance error between dynamic points. The third loss says that the estimated flows of static points should be zero. 
Finally, we assume that the flow at points from the same cluster should be consistent, and mitigate underestimation by using the proposed upper bound on the flow.
}
\label{fig:arc}
\vspace{-1.0em} 
\end{figure}
\subsection{Input and Output}
The first step to process the input data, as is commonly done, is to remove ground points from \(\mathcal P_t\) and \(\mathcal P_{t+1}\). This is typically done using HD maps~\cite{sun2020scalability,Argoverse2_2021} or ground segmentation techniques~\cite{himmelsbach2010fast}.

The estimated flow $\hat{\mathcal{F}}$ from $\mathcal{P}_t$ to $\mathcal{P}_{t+1}$ can be decomposed into two parts as follows:
\begin{equation}
    \hat{\mathcal{F}} = \mathcal{F}_{ego} + \Delta \hat{\mathcal{F}}, 
\end{equation}
where $\mathcal{F}_{ego}$ is the flow resulting from the ego vehicle's motion which can be directly obtained from $\mathbf{T}_{t,t+1}$, and $\Delta \hat{\mathcal{F}}$ is our network output.

\subsection{Model Backbone}
To enable real-time computation and estimation of scene flow across a large point set, voxelization is considered a practical encoding strategy in the model backbone. However, the reduction in resolution often leads to a poorer distinction of point features within the same voxel. With this in mind, we use DeFlow~\cite{zhang2024deflow} as the architectural basis in this paper. DeFlow integrates GRU~\cite{cho2014learning} with iterative refinement in the decoder design. 
More specifically, the GRU module maintains voxel features as its hidden state and selectively forgets or updates the information of the hidden state during each iteration according to the input point features. After multiple iterations, the optimized voxel features are concatenated with the original point features to obtain the final individual point features.
Benefiting from this design, we improve the inference efficiency compared to the commonly used backbone FastFlow3D~\cite{fastflow3d} without sacrificing the accuracy of scene flow estimation in coarse resolution settings.

\subsection{Dynamic Classification of Points}
\label{sec:dufomap}
To facilitate dynamic classification of points during the training stage, we incorporate the DUFOMap framework, a mapping-based dynamic awareness approach~\cite{daniel2024dufomap}. 
The key insight of this is that points observed inside a region that at one time has been observed as empty must be dynamic. 
Built on this insight, DUFOMap utilizes ray-casting to classify dynamic points at the sensor rate on the CPU.
The result is two disjoint sets, $\mathcal P_{t,d}$ and $\mathcal P_{t,s}$, where $\mathcal P_{t,d}$ is the set of dynamic points that have moved once inside a scene (even if they later become static) and $\mathcal P_{t,s}$ is the set of static points that did not move at any time.
Note that the dynamic-static classification is separated from the inference and training, offering our method good flexibility.

\subsection{Self-supervised Loss}
As discussed, self-supervised learning with only Chamfer distance loss is vulnerable to problems with data imbalance and incorrect associations. 
We therefore construct three additional loss functions, illustrated in~\cref{fig:arc} and described in turn below, to mitigate problems with data imbalance as well as encourage consistent object-level flow.
\subsubsection{Chamfer Distance} 
The Chamfer distance, a common self-supervised loss in existing work~\cite{mittal2020just,baur2021slim,li2021neural,li2023fast}, has the following definition:
\begin{align}
    \mathcal L_{\text{cham}} &= \frac{1}{|\hat {\mathcal P}_{t+1}|}\sum_{p \in \hat{\mathcal P}_{t+1}}\mathrm S(p, \mathcal P_{t+1}) + \frac{1}{|\mathcal P_{t+1}|}\sum_{p \in \mathcal P_{t+1}}\mathrm S(p, \hat{\mathcal P}_{t+1}) \\
    \mathrm S(p, \mathcal P_o)&=\min_{p_i \in \mathcal P_o} || p -p_i ||^2_2,
\end{align}
where, $\hat{\mathcal P}_{t+1} = \mathcal P_{t} + \hat{\mathcal{F}}_{t}$,  $\mathrm S(\cdot) = \mathrm D(\cdot)^2$, $\mathrm D(p, \mathcal P_o)$ denotes the distance between point $p$ and its nearest neighbor in $\mathcal P_o$. 

The Chamfer distance is proposed to calculate a similarity between two point clouds. However, when using it directly as a supervised signal for scene flow in autonomous driving, there are two issues we note. 
First, the number of static points is often much larger than that of dynamic points and averaging $\mathrm S(\cdot)$ over all points leads the training to favor static points, i.e., zero flow estimation.
Second, due to erroneous correspondence assumptions, only part of the flows within a dynamic object can be estimated correctly as shown in~\cref{fig:overlap_area}, where the estimated flows of points in the overlap area are zero.

To solve these issues, we contribute the following constraints using the dynamic classification information prior to supervising the network.

\subsubsection{Dynamic Chamfer Distance}
The imbalance in the number of dynamic and static points presents a significant challenge in scene flow estimation, as highlighted in previous works~\cite{zeroflow,zhang2024deflow,fastflow3d}. 
Supervised methods, provided with ground truth labels for point velocity or classification, can weight their loss functions to account for this imbalance.
For self-supervised learning, we introduce a \textit{dynamic} Chamfer distance.
In this context, `dynamic' points are identified based on the output from~\cref{sec:dufomap}. 
Specifically, this loss, \(\mathcal L_{\text{dcham}}\), is only computed on points classified as dynamic in two consecutive point clouds. By exclusively considering dynamic points, this loss captures the nuances of motion in point cloud data.
This is defined as:
\begin{equation}
    \mathcal L_{\text{dcham}} = \frac{1}{|{\hat{\mathcal P}}_{t+1,d}|}\sum_{p \in \hat{\mathcal P}_{t+1,d}}\mathrm S(p, \mathcal P_{t+1, d}) + \frac{1}{|{\mathcal P}_{t+1,d}|}\sum_{p \in \mathcal P_{t+1, d}}\mathrm S(p, \hat{\mathcal P}_{t+1,d}),
\end{equation}
where $\hat{\mathcal P}_{t+1,d} = \mathcal P_{t,d} + \hat{\mathcal{F}}_{t,d}$.

\subsubsection{Static Flow} 
The standard Chamfer distance rests on a very strong assumption: nearest neighbor matching can find perfect one-to-one correspondences of all points between two frames. 
However, due to varying observations, the number of points in the two frames differs, leading to inconsistencies and potential mismatches.
To deal with the possible mismatches in the static areas, 
we add a constraint to encourage the model to estimate zero flow for static points.
This loss is defined as follows:
\begin{equation}
    \mathcal L_\text{static} = \frac{1}{|{\mathcal P}_{t,s}|}\sum_{p \in \mathcal P_{t,s}}|| \Delta \hat {\mathcal F}(p) ||_2^2,
\end{equation}
where $\Delta \hat {\mathcal F}(p)$ represents the network output flow of point $p$.

\begin{figure}[h]
\centering
\includegraphics[trim=250 350 250 350, clip, width=\linewidth]{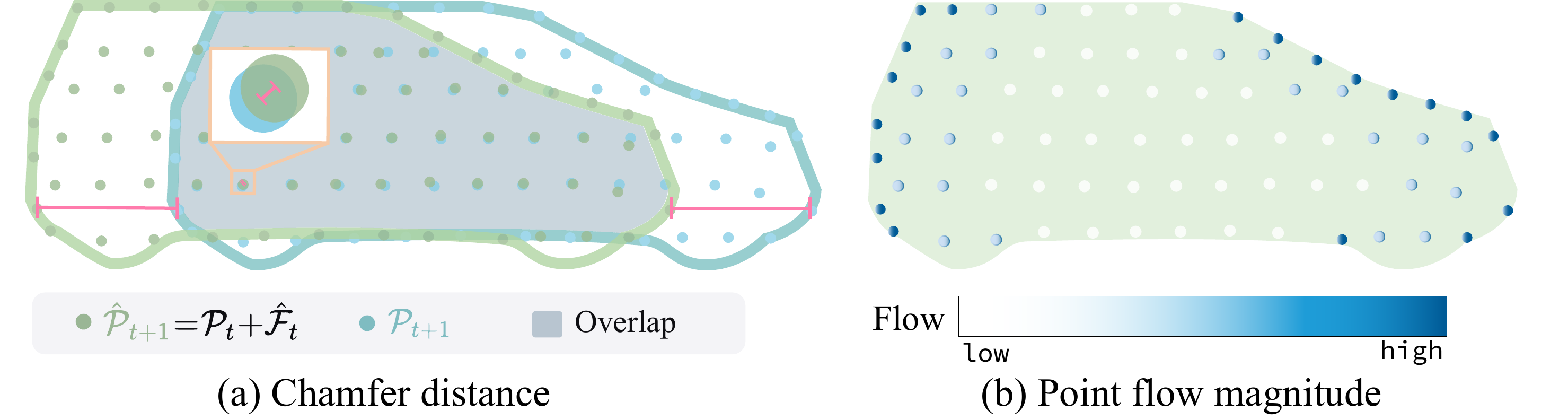}
\caption{Simple visualization of the shortcomings of using Chamfer distance as a supervisory signal for flow value estimation. The denser color on points in (b) represents higher flow values and white means the point's flow is zero. (a) illustrates how to calculate loss based on Chamfer distance, and (b) shows that the flow results, based on the nearest neighbor principle, can lead to zero flow estimation for the middle of the object.
}
\label{fig:overlap_area}
\vspace{-2.0em} 
\end{figure}

\subsubsection{Dynamic Cluster Flow} 
For dynamic fields, the inconsistencies and correspondence errors are even more complicated.
For instance, we observe that nearest neighbor matching often leads to erroneous results, e.g., on large objects undergoing translation, which is very common for moving vehicles in urban street environments as shown in \cref{fig:cover}. More specifically, \Cref{fig:overlap_area} illustrates the issue on a simplified car example. 
In this case, the flow is parallel to the object's surface, which means that nearest neighbor matching does not provide good supervision in the interior of the surface ($\mathrm D \approx 0$ for the points in the overlap area). 
This mismatch results in an incorrect local optimum that remains unresolved during the training process as it minimizes the losses based on Chamfer Distance.

Therefore, we suggest that the motion of all parts within an object (a cluster) should be approximately homogeneous over a short time interval.
We use the HDBSCAN~\cite{campello2013density,scikitlearn} clustering algorithm to identify different moving objects. This clustering is only applied to dynamic points, thus reducing the computational overhead.
\begin{equation}
    \mathcal C_{t,d} = \text{HDBSCAN} (\mathcal P_{t,d}).
\end{equation}

Directly constraining the variance of the flow distribution inside a cluster would be a straightforward idea, but does not guarantee the correctness of the obtained mean value after convergence. As we know that nearest-neighbour point correspondence can considerably underestimate flow on geometrically featureless object surfaces, we here instead propose to use an upper bound on the object motion as the supervisory signal. We derive this upper bound, per object cluster, by taking the maximum inter-frame distance of all point correspondences within the cluster.
Specifically, we exploit information from the original input point cloud data, i.e., $\mathcal P_t$ and $\mathcal P_{t+1}$. 
After the dynamic classification and clustering process, we find the index of the point $p_k$ in cluster $c_i \in \mathcal C_{t,d}$ with the largest distance to its nearest neighbor point in $\mathcal{P}_{t+1,d}$, i.e.,  
\begin{equation}    
    \kappa_i = \argmax_k \{\mathrm D(p_k,\mathcal P_{t+1,d}) | p_k \in \mathcal P_{c_i}\}. \label{eq:k}
\end{equation}
We calculate the upper bound, $\tilde f_{c_i}$ on the flow for cluster $c_i$ as 
\begin{equation}
    \tilde f_{c_i} = p'_{\kappa_i} - p_{\kappa_i},
    \label{eq:f_ci}
\end{equation}
where $p'_{\kappa_i}$ is the nearest neighbor of point $p_{\kappa_i}$ in $\mathcal P_{t+1,d}$.
We use this to drive the estimated flows of cluster $c_i$ towards $\tilde f_{c_i}$ as follows:
\begin{align}
    \mathcal L_{c_i}&=\sum_{p_j \in \mathcal P_{c_i}} || \hat f_{p_j} - \tilde f_{c_i} ||^2_2, \\
    \mathcal L_{\text{dcls}}&=\frac{1}{|\mathcal{P}_{t,d}|}{\sum_{c_i \in \mathcal {C}_{t,d}} \mathcal L_{c_i}}.
\end{align}

In conclusion, the final  SeFlow loss incorporates all four losses introduced above, 
\begin{equation}
    \mathcal L_{total} = \mathcal L_{\text{cham}} + \mathcal L_\text{static} + \mathcal L_{\text{dcham}} + \mathcal L_{\text{dcls}}.
\end{equation}
The effect of each loss will be evaluated in the ablation study in \cref{sec:ablation_study}.

\section{Experiment}
\label{sec:exp}
In this section, we first outline evaluation details and then present quantitative comparisons with state-of-the-art methods on two benchmark datasets. 
A series of ablation studies are presented to better evaluate the individual components of our approach. Finally, we conclude with qualitative results on Argoverse 2 and discuss limitations of the approach.

\subsection{Dataset and Metric}
We briefly introduce the dataset and metrics for evaluation in the following section. More implementation details, such as hyperparameters for training and dataset descriptions, can be found in the supplementary material.

\subsubsection{Dataset} 
We evaluate our approach on two large-scale autonomous driving datasets: Argoverse 2~\cite{Argoverse2_2021} and Waymo~\cite{sun2020scalability}. 
Ground removal is performed using HD maps for both datasets as described in~\cite{Argoverse2_2021}. Waymo datasets contain 798 training and 202 validation scenes respectively. 
We focus our description here on Argoverse 2, which provides official and public scene flow challenges~\cite{onlineleaderboard,onlineleaderboard2}, with the \textit{Sensor} and \textit{LiDAR} datasets. 
The \textit{Sensor} dataset encompasses 700 training and 150 validation scenes. Each scene is approximately 15 seconds long in 10$\mathrm{~Hz}$, with complete annotations for evaluation. 
The evaluation for another 150 test scenes can be accessed indirectly by submitting a solution to the leaderboard.
The \textit{LiDAR} dataset contains 20,000 scenes without any annotation and is only used as extra data in~\cref{sec:ab_datasize}.

\subsubsection{Metric}
The benchmark follows existing works~\cite{zhang2024deflow,zeroflow,chodosh2023re,onlineleaderboard} and uses the three-way End Point Error.
End Point Error (EPE), as defined in~\cref{eq:opti}, measures the L2 norm of the discrepancy between the predicted and actual flow vectors, expressed in meters. 
The EPE three-way (3-way) is defined as the unweighted average EPE across three disjoint sets of points: Foreground Dynamic (FD), Foreground Static (FS), and Background Static (BS). 
The definition of `dynamic' is as follows: If the flow at a point exceeds the threshold (the public leaderboard setting is 0.05m), the point is defined as dynamic. 
Given the 10 Hz sensor frequency, this threshold corresponds to a speed of $0.5 \mathrm{~m} / \mathrm{s}$. All evaluations are limited to points that are within a 100m \(\times\) 100m box centered on the ego vehicle. 

\begin{table}[t!]
\centering
\def\arraystretch{1.3}
\caption{
Comparisons on Argoverse 2 \underline{test set} from the online leaderboard~\cite{onlineleaderboard}. 
Upper groups are supervised methods, lower groups are \textbf{self-supervised} methods. 
Our method achieves state-of-art performance in the self-supervised scene flow task. $^\dagger$ means methods can run in real-time (10 Hz) onboard.
}
\setlength{\tabcolsep}{1.5mm}{
\begin{tabular}{lccccc} 
\toprule
\multirow{2}{*}{Method} & \multirow{2}{*}{\begin{tabular}[c]{@{}c@{}}Run Time\\per frame [ms]\end{tabular}} & \multicolumn{4}{c}{EPE  ↓} \\ 
\hhline{~~----}
& & \multicolumn{1}{c}{{\cellcolor[rgb]{0.949,0.949,0.949}}3-way} & \multicolumn{1}{c}{FD} & \multicolumn{1}{c}{FS} & \multicolumn{1}{c}{BS}                                                                       \\ 
\hline
FastFlow3D$^\dagger$~\cite{fastflow3d}       & 34 $\pm$ 5             & {\cellcolor[rgb]{0.949,0.949,0.949}}0.0782              & 0.2072      & 0.0253       & 0.0020  \\
DeFlow$^\dagger$~\cite{zhang2024deflow}      & 48 $\pm$ 4             & {\cellcolor[rgb]{0.949,0.949,0.949}}0.0534              & 0.1340      & 0.0232       & 0.0029  \\ 
\hline                                                                                                                                                                   
FastNSF~\cite{li2023fast}                    & 507 $\pm$ 312          & {\cellcolor[rgb]{0.949,0.949,0.949}}0.1657              & 0.3540      & 0.0406       & 0.1025  \\
NSFP~\cite{li2021neural}                     & 32,060 $\pm$ 10,112    & {\cellcolor[rgb]{0.949,0.949,0.949}}0.0685              & 0.1503      & 0.0302       & 0.0248  \\
ZeroFlow$^\dagger$~\cite{zeroflow}           & 34 $\pm$ 5             & {\cellcolor[rgb]{0.949,0.949,0.949}}0.0814              & 0.2109      & 0.0254       & 0.0080  \\
SeFlow (Ours)$^\dagger$                      & 48 $\pm$ 4             & {\cellcolor[rgb]{0.949,0.949,0.949}}\textbf{0.0628}     & 0.1525      & 0.0321       & 0.0038  \\
\bottomrule
\end{tabular}}
\vspace{-0.6em}
\label{tab:main_res}
\end{table}

\subsection{Quantitative Results}
\label{sec:quan}
We evaluate the performance of our method SeFlow and compare it with the currently best performing methods on the Argoverse 2 test set and Waymo validation set. In \cref{tab:main_res} and \cref{tab:waymo_res}, the upper group, FastFlow3D and DeFlow, are supervised methods trained with ground truth flow.
Compared to the supervised methods, we note that SeFlow outperforms FastFlow3D and approaches the level of DeFlow in terms of EPE 3-way. 
This comparative analysis demonstrates the great potential of self-supervised learning in scene flow estimation and validates the effectiveness of our approach.

In the self-supervised category, our SeFlow method achieves state-of-the-art performance on both datasets. While FastNSF and NSFP are optimization-based methods that do not rely on pre-trained weights for subsequent estimations, their inference times are not conducive to real-time requirements. NSFP, with the second best result in~\cref{tab:main_res}, takes approximately 30 seconds to predict a single frame, which is impractical for real-time autonomous driving applications. FastNSF, on the other hand, enhances efficiency through voxelization for quicker Chamfer distance calculation. 
Although significantly faster than NSFP, the performance of FastNSFP is the worst among all methods evaluated.

ZeroFlow is capable of estimating scene flow in real time, with an accuracy similar to NSFP (slightly better on Waymo and slightly worse on Argoverse 2). SeFlow stands out not only for achieving the best result, but also for drastically reducing the run time to 50 milliseconds, which is more than two orders of magnitude faster than NSFP while providing more accurate flow estimates. 
SeFlow's state-of-the-art performance demonstrated in \cref{tab:main_res} and \cref{tab:waymo_res} underscores the effectiveness of our novel self-supervised method in scene flow estimation.
\begin{table}[t!]
\centering
\def\arraystretch{1.3}
\caption{
Comparisons on Waymo Open dataset \underline{validation set}. Upper groups are supervised methods, lower groups are \textbf{self-supervised} methods. 
Our method achieves state-of-art performance in the self-supervised scene flow task. $^\dagger$ means methods can run in real-time (10 Hz) onboard.
}
\setlength{\tabcolsep}{1.5mm}{
\begin{tabular}{lccccc} 
\toprule
\multirow{2}{*}{Method} & \multirow{2}{*}{\begin{tabular}[c]{@{}c@{}}Run Time\\per frame [ms]\end{tabular}} & \multicolumn{4}{c}{EPE  ↓} \\ 
\hhline{~~----}
& & \multicolumn{1}{c}{{\cellcolor[rgb]{0.949,0.949,0.949}}3-way} & \multicolumn{1}{c}{FD} & \multicolumn{1}{c}{FS} & \multicolumn{1}{c}{BS}                                                                     \\ 
\hline
FastFlow3D$^\dagger$~\cite{fastflow3d}       & 27 $\pm$ 6             & {\cellcolor[rgb]{0.949,0.949,0.949}}0.0782              & 0.1954      & 0.0246       & 0.0152  \\
DeFlow$^\dagger$~\cite{zhang2024deflow}      & 42 $\pm$ 4             & {\cellcolor[rgb]{0.949,0.949,0.949}}0.0446              & 0.0980      & 0.0259       & 0.0098  \\
\hline                                                                                                                                                                   
FastNSF~\cite{li2021neural}              & 593 $\pm$ 308           & {\cellcolor[rgb]{0.949,0.949,0.949}}0.1579              & 0.3012      & 0.0146       & 0.0403  \\
NSFP~\cite{li2023fast}                    & 76,163 $\pm$ 32,256          & {\cellcolor[rgb]{0.949,0.949,0.949}}0.1005              & 0.1712      & 0.1081       & 0.0221  \\
ZeroFlow$^\dagger$~\cite{zeroflow}           & 27 $\pm$ 6             & {\cellcolor[rgb]{0.949,0.949,0.949}}0.0921              & 0.2162      & 0.0153       & 0.0241  \\
SeFlow (Ours)$^\dagger$         & 42 $\pm$ 4             & {\cellcolor[rgb]{0.949,0.949,0.949}}\textbf{0.0598}     & 0.1506      & 0.0181       & 0.0106  \\
\bottomrule
\end{tabular}}
\vspace{-0.6em}
\label{tab:waymo_res}
\end{table}

\subsection{Ablation study}
\label{sec:ablation_study}
In this section, we delve into two key aspects of our SeFlow pipeline. Firstly, we examine the impact of different loss terms on the accuracy of our flow prediction results. This analysis aims to demonstrate the necessity and effectiveness of the loss components we propose. Secondly, we explore how the size of the training dataset, especially in the case of limited training data, affects the outcomes of our self-supervised training process.

\subsubsection{Loss Terms}
The advantages of each loss design are evident in \cref{tab:loss} evaluated on the Argoverse 2 validation set with 20 training epochs. Instead of relying solely on the chamfer loss $\mathcal L_{\text{cham}}$, we investigate how incorporating the additional three losses ($\mathcal L_{\text{static}}$, $\mathcal L_{\text{dcham}}$, $\mathcal L_{\text{dcls}}$) can boost the performance. 

\begin{table}[t!]
\centering
\def\arraystretch{1.3}
\caption{
Ablation study of loss terms. Results are evaluated on the Argoverse 2 \underline{validation set} with 20 training epochs.}
\setlength{\tabcolsep}{1mm}{
\begin{tabular}{ccccc|cccc} 
\toprule
\multirow{2}{*}{Exp. Id} & \multirow{2}{*}{$\mathcal L_{\text{cham}}$} & \multirow{2}{*}{$\mathcal L_{\text{dcham}}$} & \multirow{2}{*}{$\mathcal L_{\text{static}}$} & \multirow{2}{*}{$\mathcal L_{\text{dcls}}$} & \multicolumn{4}{c}{EPE ↓}     \\ 
\hhline{~~~~~----}
& &  &  & & \multicolumn{1}{c}{{\cellcolor[rgb]{0.949,0.949,0.949}}3-way} & \multicolumn{1}{c}{FD} & \multicolumn{1}{c}{FS} & \multicolumn{1}{c}{BS}  \\ 
\hline
1 & \checkmark &            &            &            & {\cellcolor[rgb]{0.949,0.949,0.949}0.0962}          & 0.203             & 0.052             & 0.033\\
2 & \checkmark & \checkmark &            &            & {\cellcolor[rgb]{0.949,0.949,0.949}0.0916}          & 0.181             & 0.059             & 0.035\\
3 & \checkmark & \checkmark & \checkmark &            & {\cellcolor[rgb]{0.949,0.949,0.949}0.0779}          & 0.220             & \textbf{0.012}    & \textbf{0.002}\\
4 & \checkmark & \checkmark & \checkmark & \checkmark & {\cellcolor[rgb]{0.949,0.949,0.949}\textbf{0.0643}} & \textbf{0.160}    & 0.029             & 0.004  \\
\bottomrule
\end{tabular}
}
\vspace{-0.6em}
\label{tab:loss}
\end{table}

After adding the dynamic Chamfer loss \(\mathcal L_{\text{dcham}}\), experiment 2 shows a decrease in flow estimation error of about 0.22 (10\%) for foreground dynamics (FD), while the static errors (FS, BS) remain essentially the same. 
Experiment 3 then incorporates \(\mathcal L_{\text{static}}\) constraint, which significantly reduces the foreground and background static flow estimation errors, 80\% for FS and 94\% for BS.
However, adding \(\mathcal L_{\text{static}}\) also increases the foreground dynamic error. We attribute this to the difficulty of estimating the flow of moving, geometrically featureless, objects as mentioned previously, which would be reinforced by  \(\mathcal L_{\text{static}}\). 
Even so, considering the two static components in the EPE 3-way metric, the EPE 3-way would still benefit considerably from this loss (15\% error reduction).
Finally, in experiment 4, we incorporate the \(\mathcal L_{\text{dcls}}\). This results in a notable decrease in overall EPE 3-way by 33\% compared to solely using the chamfer loss (experiment 1). 
The above experiments illustrate that our method is not a simple stack of losses, but a complementary holistic design.
\vspace{-0.6em}

\subsubsection{Training Dataset Size}
\label{sec:ab_datasize}
In the context of robotics and autonomous driving, there are situations where the number of accessible frames is limited.
This section evaluates the performance of SeFlow given different amounts of training data.

In this experiment, we denote the entire Argoverse 2 \textit{Sensor} training set as 100\%.
Extra unlabeled data can be retrieved from the \textit{LiDAR} dataset for self-supervised methods.
To assess the impact of dataset size, we conduct the same 50 epochs of training for methods using 10\%, 20\%, 50\%, and 100\% of the total data in~\cref{tab:datasize} and~\cref{fig:datasize} and evaluate the resulting models using the validation set.

From~\cref{tab:datasize}, we can observe that with only 20\% or 50\% training data, our SeFlow can already outperform ZeroFlow which uses 100\% or 200\% data. \Cref{fig:datasize} illustrates even more intuitively that our SeFlow can easily outperform both existing supervised (FastFlow3D) and unsupervised (ZeroFlow) approaches with the same amount of data. 
We attribute the demonstrated data efficiency of our method to the well-designed loss functions, which integrate a dynamic awareness mapping method into our framework and enable better scene-level comprehension.

\vspace{2.0em}
\noindent
\begin{minipage}{0.5\textwidth}
\centering
\def\arraystretch{1.3}

\setlength{\tabcolsep}{1.0mm}{
\begin{tabular}{ccccc}
\toprule
\multirow{2}{*}{Dataset} 
& \multicolumn{4}{c}{EPE ↓}  \\
\hhline{~----}
& \multicolumn{1}{c}{{\cellcolor[rgb]{0.949,0.949,0.949}}3-way} & \multicolumn{1}{c}{FD} & \multicolumn{1}{c}{FS} & \multicolumn{1}{c}{BS}\\ 
\hline
10\%        & {\cellcolor[rgb]{0.949,0.949,0.949}}0.094          & 0.234               & 0.040        & 0.006           \\
20\%        & {\cellcolor[rgb]{0.949,0.949,0.949}}0.078          & 0.197               & 0.032        & 0.004           \\
50\%        & {\cellcolor[rgb]{0.949,0.949,0.949}}0.066          & 0.167               & 0.028        & 0.004           \\
100\%       & {\cellcolor[rgb]{0.949,0.949,0.949}}\textbf{0.059} & 0.147               & 0.026        & 0.004           \\
\hline
ZF 100\% & {\cellcolor[rgb]{0.949,0.949,0.949}}0.088             & 0.231               & 0.022        & 0.011         \\
ZF 200\% & {\cellcolor[rgb]{0.949,0.949,0.949}}0.076 & 0.198 & 0.018 & 0.011  \\
\bottomrule
\end{tabular}}
\captionof{table}{
Ablation study of dataset size compared with Zeroflow (ZF), another self-supervised learning method. Results are evaluated on the Argoverse 2 \underline{validation set} with 50 training epochs. The total dataset (100\%) is 107k frames. 
200\% data means all of the \textit{Sensor} dataset plus an equal amount of the \textit{LiDAR} dataset (214k frames). 
}
\label{tab:datasize}
\end{minipage}
\hspace{1em}
\begin{minipage}{0.46\textwidth}
  \centering
  \includegraphics[width=0.95\linewidth]{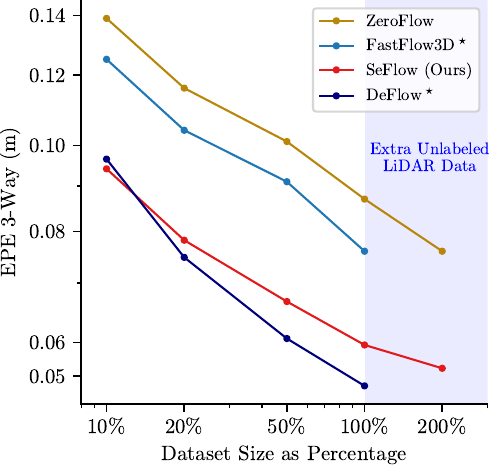}
  \hspace{-2em}\captionof{figure}{The relationship between flow estimation error and training dataset size, scaled in \(\log_{10}\). Methods with $^\star$ are supervised by ground truth labels. SeFlow uses less data but gets comparable results compared to FastFlow3D and ZeroFlow. }
  \label{fig:datasize}
\end{minipage}
\vspace{-0.8em}

\subsection{Qualitative Results} 
\begin{figure}[h]
\centering
\includegraphics[trim=15 20 15 90, clip, width=\linewidth]{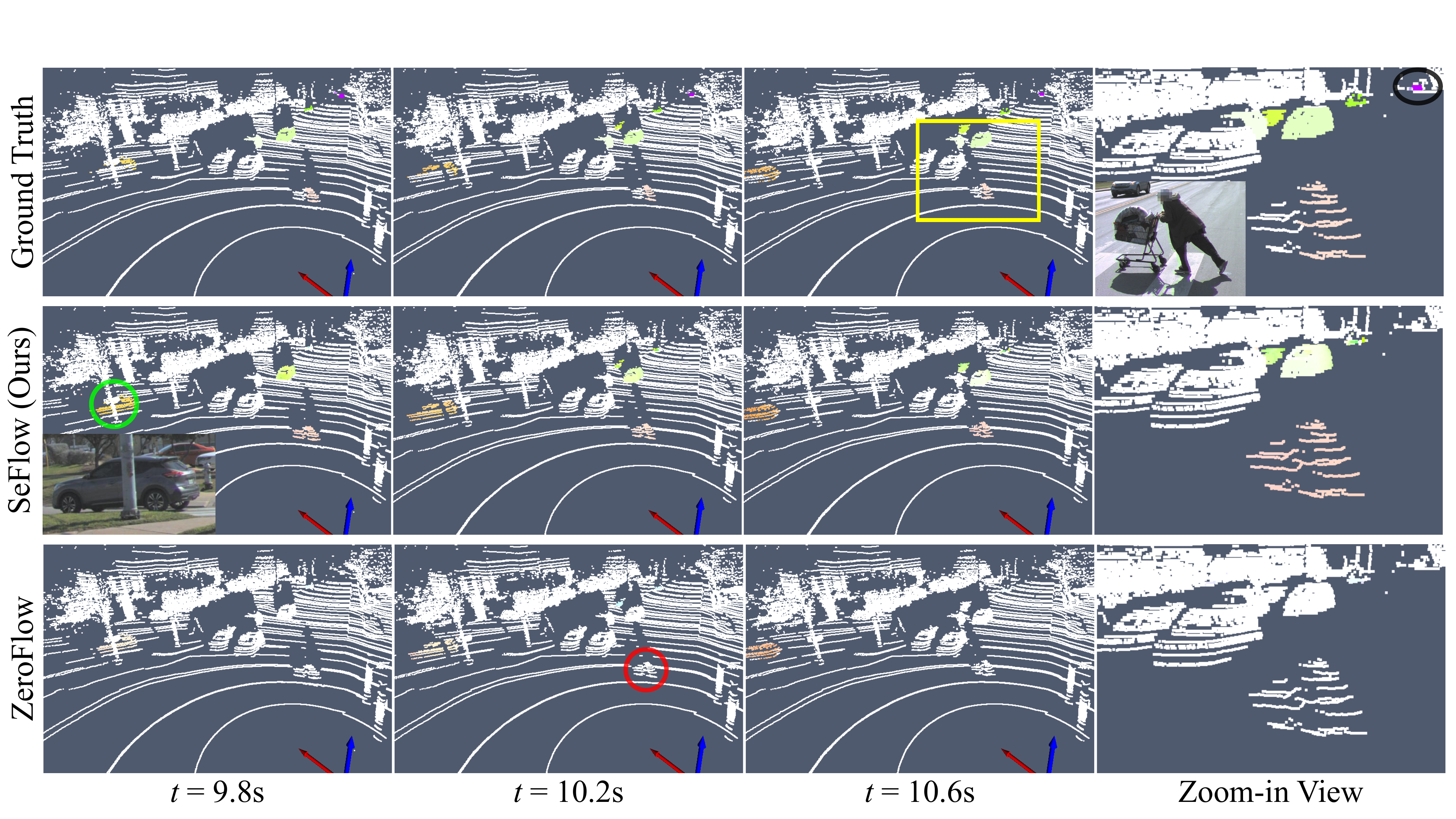}
\caption{Qualitative results from Argoverse 2 validation set. The top row displays the ground truth flow, the middle row presents the SeFlow result, and the bottom row showcases another self-supervised method ZeroFlow\cite{zeroflow} result. Different color indicates different directions and more saturated color means larger flow estimation. Ego motion is compensated for a clearer view.
}
\label{fig:q1}
\vspace{-1.0em} 
\end{figure}

\Cref{fig:q1} presents a qualitative flow estimation result on a sequence of scenes in the Argoverse 2 validation set. 
More visualization can be found in the supplementary material.
The first three columns in \cref{fig:q1} showcase the same scene at different timestamps, and the last column shows a zoomed-in view of the scene.

In the first column featuring a vehicle partially occluded by a traffic pole, SeFlow demonstrates superior flow estimation capabilities compared to ZeroFlow. 
Sometimes, SeFlow can even detect flow overlooked by the ground truth annotations. Argoverse 2 derives ground truth flow from manual instance level labels, and as a result, the flow of points outside the bounding box may be ignored. 
This issue is particularly pronounced in smaller objects, and an example of this can be seen in the fourth column where we zoom in on a case. 
In this example, a pedestrian is pushing a shopping cart across the road. By comparing the first and the third column, we can observe that both objects are moving. However, the ground truth labels zero flow (white) for the shopping cart. SeFlow, on the other hand, successfully predicts consistent flow even at this small scale.
In comparison, the baseline method ZeroFlow suffers at flow prediction of small objects and regards both the pedestrian and the shopping cart as static (white).

Although our method shows superior scene flow estimation compared to other baseline methods, it also has some limitations.
Based on the zoom-in view in the fourth column in \cref{fig:q1}, we can find some purple flow labels at the top right corner in the first row (Ground Truth). Both SeFlow (Ours) and ZeroFlow fail to predict any flow for these points. 
One possible reason is that the point cloud of distant objects is sparse and easily ignored by both the voxelization process and clustering algorithms.
This makes it difficult to estimate the flow of distant objects using only two consecutive frames. 
Based on this, multi-modality or time-consistent networks would be a further direction for future research.

\section{Conclusions}
In this paper, we proposed SeFlow, an efficient and effective self-supervised training method for scene flow estimation in autonomous driving with large-scale point clouds as input. 
Our primary contributions include a learning method that integrates static and dynamic awareness to construct self-supervision objectives. 
We further identify problems with the correspondence assumptions of Chamfer-based loss functions commonly used for self-supervised learning, and mitigate these with a constraint based on the upper bound of object motion.
Our experimental results underscore the efficacy of our approach.

Future research may concentrate on integrating multi-modality (e.g., cameras and radar) within the SeFlow framework for higher flow estimation accuracy or embedding temporal coherence concepts within our pipeline. Additionally, the optimization of multi-loss learning weights warrants further exploration.

\ifready
    \section*{Acknowledgement}
    Thanks to RPL member: Li Ling helps review this manuscript. Thanks to Kyle Vedder (the ZeroFlow paper author), who kindly opened his code including pre-trained weights, and discussed their result with us which helped this work a lot. 
    We also thank the anonymous reviewers for their constructive comments.
    This work was partially supported by the Wallenberg AI, Autonomous Systems and Software Program (WASP) funded by the Knut and Alice Wallenberg Foundation and Prosense (2020-02963) funded by Vinnova. The computations were enabled by the supercomputing resource Berzelius provided by National Supercomputer Centre at Linköping University and the Knut and Alice Wallenberg Foundation, Sweden.
\fi

%
%
\bibliographystyle{splncs04}
\bibliography{ref}

\newpage

\appendix

\begin{center}
    {\textbf{\Large {SeFlow Supplementary Material}}}\\
\end{center}

In this supplementary material, we begin by detailing the implementation aspects, including datasets and hyperparameters for dynamic classification, clustering, and training, as outlined in \cref{sec:detail} and \cref{sec:data}. 
Following this, we present additional results in several key areas:
\begin{itemize}
    \item \Cref{sec:loss} (Loss Terms): This section explores various applications of static and dynamic classification in designing loss functions. To validate the effectiveness of our proposed upper bound flow in the object cluster, we also incorporate common strategies such as averaging or maximizing the estimated flows within clusters. Furthermore, an additional ablation study table is provided to further elucidate the impact of our loss terms.
    \item \Cref{sec:model_backbone} (Different Model Backbones): This section demonstrates that SeFlow's effectiveness is not limited to a specific model backbone. We show that also with the same backbone as FastFlow3D, SeFlow outperforms both self-supervised and supervised baselines, underscoring the strength of our self-supervised pipeline.
    \item \Cref{sec:qualitative} (Qualitative Results): In addition to the sequence scenes discussed qualitatively in the main paper, we present two more qualitative results showcasing SeFlow's performance on both Argoverse 2 and Waymo datasets, including some failure cases.
\end{itemize}

\section{Experiment}

\subsection{Implementation Details}
\label{sec:detail}

\subsubsection{Our Method} The resolution of network voxelization is set as $0.2\mathrm{~m}$, consquently, the $[512,512]$ grid corresponds to a $102.4\mathrm{~m}\times 102.4\mathrm{~m}$ map. 
DUFOMap~\cite{daniel2024dufomap} is used for the dynamic classification in our method and its resolution is consistent with the voxelization setting of $0.2\mathrm{~m}$. For DUFOMap's parameters $d_p$ and $d_s$, we keep them as default which is 1 and 0.2 respectively. 
HDBSCAN only clusters dynamic points inside $\mathcal{P}_{t,d}$ where the minimum cluster size is set to 20 and cluster selection $\epsilon$ is set to 0.7.
Our SeFlow is trained for 50 epochs with a batch size of 80 without any ground truth labels. We employ Adam optimizer with a $2\times 10^{-6}$ learning rate. 
The code is open-sourced at \url{https://github.com/KTH-RPL/SeFlow}. 
The latest leaderboard \cite{onlineleaderboard2} result model is trained for 15 epochs with a total batch size of 64 and $2\times 10^{-4}$ learning rate on four A100 GPUs for around 10 hours.

\subsubsection{Other Methods}
In our main comparison on the Argoverse 2 test set (as shown in Table 1 of our main paper), we directly reference results from the online leaderboard~\cite{onlineleaderboard} and Table I in DeFlow~\cite{zhang2024deflow}, with result files available in their discussion thread\footnote{\url{https://github.com/KTH-RPL/DeFlow/discussions/2}}.
For the Waymo validation set results (Table 2 in our main paper), we present the outcomes for FastFlow3D~\cite{fastflow3d}, ZeroFlow~\cite{zeroflow}, and NSFP~\cite{li2021neural} as reported in Table 2 of ZeroFlow~\cite{zeroflow}, which were trained on the Waymo train set. FastNSF~\cite{li2023fast} results were obtained by running their model\footnote{\url{https://github.com/Lilac-Lee/FastNSF}} with default Waymo settings. For DeFlow~\cite{zhang2024deflow} and our method SeFlow, we conducted our own training on the Waymo training set, adhering to the same training strategy outlined earlier.
Regarding all ZeroFlow entries in our ablation study, we utilized the pre-trained weights available in their official repository\footnote{\url{https://github.com/kylevedder/zeroflow_weights}}.

\subsubsection{Setup} For inference, to measure the complexity and computational cost of our model and other methods, all experiments are executed on a desktop powered by an Intel Core i9-12900KF CPU and equipped with a GeForce RTX 3090 GPU.

\subsubsection{Process Time}
Regarding the processing time on labeling DUFOMap and HDBSCAN steps, taking the Argoverse 2 dataset as an example, the average runtimes of DUFOMap and HDBSCAN are 50ms/frame and 500ms/frame respectively. It takes approximately 16.35 CPU hours for the whole dataset in the setup we mentioned above. 
The DUFOMap and HDBSCAN steps are pre-processed before training to avoid redundant computations. 
SeFlow's training time compared to DeFlow is 30 hours vs 21 hours on 8 A100 GPUs  with a $2\times 10^{-6}$ learning rate. 
Note that for the result on the latest leaderboard \cite{onlineleaderboard2}, we enlarge the learning rate and smaller the total epoch result model is trained for 15 epochs with a total batch size of 64 and $2\times 10^{-4}$ learning rate on four A100 GPUs for around \textit{10 hours}.

\subsection{Datasets}
In this section, we present the datasets we use. For the convenience of the reader and to make these presentations self-contained there are some repetitions from the main paper.
\label{sec:data}

\subsubsection{Argoverse 2~\cite{Argoverse2_2021}} It contains two subdataset - \textit{Sensor} and \textit{Lidar}.
The \textit{Sensor} dataset encompasses 700 training and 150 validation scenes. Each scene is approximately 15 seconds long in 10$\mathrm{~Hz}$, complete with annotations for evaluation. 
The \textit{LiDAR} dataset lacks imagery and any other annotations containing 16,000 training, 2,000 validation, and 2,000 test scenes, respectively. Each scene is approximately 30 seconds long in 10$\mathrm{~Hz}$. The \textit{LiDAR} dataset is designed to support research into self-supervised learning in the lidar domain, as well as point cloud forecasting. 
All of the above datasets are collected using two 32-channel LiDARs. The average number of points in one frame is around 52,871 after ground removal. 
The 200\% data (214k frames in total) in the main paper means 100\% \textit{Sensor} dataset which contains 107k frames plus another 107k frames from \textit{LiDAR} dataset, selected via the same process as in \cite{zeroflow}. 
\cite{zeroflow} uniformly sampled 12 pairs of frames from each scene of the \textit{LiDAR} dataset first, followed by random sampling to get another 107k frames, i.e., another 100\% of data.

\subsubsection{Waymo Open Dataset~\cite{sun2020scalability}}
The dataset contains 798 training and 202 validation sequences. Each sequence contains 20 seconds of 10Hz point clouds collected using a custom LiDAR mounted on the roof of a car. The total number of training frames is 155k. The average number of points in one frame is around 79,327 after ground removal. 
Since it does not have a public leaderboard or official evaluation scripts, in this paper, we follow the same setting and process steps as ZeroFlow~\cite{zeroflow} to make fair comparisons. The evaluation follows the same Argoverse 2 official evaluation scripts and evaluates flow performance on points that do not belong to the ground and are within a 100m \(\times\) 100m range centered on the origin.

\subsection{Additional Ablation Studies in Loss Terms}
\label{sec:loss}

\subsubsection{Dynamic Chamfer and Static Loss}
We investigate different alternatives for the design of dynamic Chamfer and static losses. In addition to the standard Chamfer loss $\mathcal L_{\text{cham}}$, both FastFlow3D \cite{fastflow3d} and DeFlow \cite{zhang2024deflow} propose different losses and weights for static and dynamic points based on ground truth classification labels. In the comparison, we reformat the dynamic weight loss formulas of these two methods, making use of the classification results:

\begin{align}
    \text{\cite{fastflow3d}}: \mathcal{L}_{d,s} &= \frac{1}{|\mathcal{P}_t|} \sum_{p \in \mathcal{P}_t} \sigma(p)\mathrm S(p), \text { where }\sigma(p) = \begin{cases}0.9 & \text { if } p \in \mathcal P_{t,d} \\ 0.1 & \text { if } p \in \mathcal P_{t,s}\end{cases}\label{eq:scaled_ds}.\\
    \text{\cite{zhang2024deflow}}: \mathcal{L}_{d,s} &= \frac{1}{|\mathcal{P}_{t,d}|} \sum_{p \in \mathcal{P}_{t,d}} \mathrm S(p) + \frac{1}{|\mathcal{P}_{t,s}|} \sum_{p \in \mathcal{P}_{t,s}} \mathrm S(p)\label{eq:unweight_ds}.
\end{align}
In the above formulas, $\mathrm S(\cdot) = \mathrm D(\cdot)^2$, and $\mathrm D(p, \mathcal P_{t+1})$ denotes the distance between point $p$ and its nearest neighbor in $\mathcal P_{t+1}$. 
Since we do not use any labels, the dynamic points $\mathcal{P}_{t,d}$ and static points $\mathcal{P}_{t,s}$ in ~\cref{eq:scaled_ds} (FastFlow3D strategy) and ~\cref{eq:unweight_ds} (DeFlow strategy) are obtained from the dynamic classification results.
As a comparison, our dynamic and static losses can be represented as:
\begin{equation}
    \mathcal{L}_{d,s} = \mathcal L_{\text{dcham}}+\mathcal L_{\text{static}}\label{eq:our_ds}.
\end{equation}

\begin{table}[h]
\centering
\def\arraystretch{1.3}
\caption{
Ablation study on different static dynamic usage in loss design. Our design is $\mathcal L_{\text{cham}} + \mathcal L_{\text{dcham}} + \mathcal L_{\text{static}}$, while others are $\mathcal L_{\text{cham}} + \mathcal L_{d,s}$.
}
\setlength{\tabcolsep}{1.5mm}{
\begin{tabular}{ccccc} 
\toprule
\multirow{2}{*}{Stategy} & \multicolumn{4}{c}{EPE}\\ 
\cline{2-5}
& \multicolumn{1}{c}{{\cellcolor[rgb]{0.949,0.949,0.949}3-way}} & \multicolumn{1}{c}{FD} & \multicolumn{1}{c}{FS} & \multicolumn{1}{c}{BS}  \\ 
\hline
\cref{eq:scaled_ds} \cite{fastflow3d}   & {\cellcolor[rgb]{0.949,0.949,0.949}0.094} & \textbf{0.192} & 0.057 & 0.034 \\
\cref{eq:unweight_ds} \cite{zhang2024deflow} & {\cellcolor[rgb]{0.949,0.949,0.949}0.099} & 0.211 & 0.053 & 0.033  \\
\cref{eq:our_ds} Ours & {\cellcolor[rgb]{0.949,0.949,0.949}\textbf{0.078}} & 0.220    & \textbf{0.012}     & \textbf{0.002}  \\
\bottomrule
\end{tabular}}
\label{tab:ab_static_dynamic}
\end{table}

\Cref{tab:ab_static_dynamic} shows that under the self-supervised strategy, our static and dynamic loss design with $\mathcal L_{\text{dcham}}+\mathcal L_{\text{static}}$ is the best solution according to EPE 3-way. 
Looking at the individual EPE components, EPE FD is similar to the three strategies and the main difference is in the static EPE components (FS and BS) where our strategy results in significantly lower errors.
We attribute this to targeted loss selection rather than just loss weight balancing.

\subsubsection{Cluster Loss}
To show the superiority of our cluster loss design, we experimented with different designs to determine \(\tilde f_{c_i}\). 
\Cref{tab:loss_design} presents a comparison of different cluster flow loss configurations under the following definitions: 
\begin{align}
    \text{avg}: \tilde f_{c_i} &= \frac{1}{|\mathcal P_{c_i}|}\sum_{p \in \mathcal P_{c_i}}{\hat{\mathcal F}(p)}. \label{eq:f_mean} \\
    \text{max}: \tilde f_{c_i} &= \max_{p \in \mathcal P_{c_i}}\hat{\mathcal F}(p).
    \label{eq:f_max} \\
    \text{Ours}: \tilde f_{c_i} &= p'_{\kappa} - p_{\kappa}, 
    \text{ where }\kappa = \argmax \{\mathrm D(p_k,\mathcal P_{t+1,d}) | p_k \in \mathcal P_{c_i}\}.\label{eq:f_ci}
\end{align}
In the above formulas, $\hat{\mathcal{F}}(p)$ represents the estimated flow of point $p$ and $p'_{\kappa}$ is the nearest neighbor of $p_{\kappa}$ in $\mathcal P_{t+1,d}$.
We explored the average of the estimated flow (\cref{eq:f_mean}), the maximum from the estimated flow (\cref{eq:f_max}), and our proposed method as detailed in~\cref{eq:f_ci}.
\begin{table}[h]
\centering
\def\arraystretch{1.3}
\caption{
Ablation study on different cluster flow consistency. All variations utilize four losses, and the only difference is the choice of $\tilde f_{c_i}$.
}
\setlength{\tabcolsep}{1.5mm}{
\begin{tabular}{ccccc} 
\toprule
\multirow{2}{*}{$\tilde f_{c_i}$ } & \multicolumn{4}{c}{EPE}\\ 
\cline{2-5}
& \multicolumn{1}{c}{{\cellcolor[rgb]{0.949,0.949,0.949}3-way}} & \multicolumn{1}{c}{FD} & \multicolumn{1}{c}{FS} & \multicolumn{1}{c}{BS}  \\ 
\hline
\cref{eq:f_mean} $\text{avg}$ & {\cellcolor[rgb]{0.949,0.949,0.949}0.078} & 0.221  & \textbf{0.012}     & 0.002  \\
\cref{eq:f_max} $\max$      & {\cellcolor[rgb]{0.949,0.949,0.949}0.092} & 0.262    & 0.013     & \textbf{0.001}  \\
\cref{eq:f_ci} Ours         & {\cellcolor[rgb]{0.949,0.949,0.949}\textbf{0.064}} & \textbf{0.160}     & 0.029    & 0.004    \\
\bottomrule
\end{tabular}}
\label{tab:loss_design}
\end{table}

The results in~\cref{tab:loss_design} demonstrate that our method decreases the EPE of foreground dynamics the most among the three definitions, which contributes significantly to the reduction of 3-way EPE. 
Compared to the huge improvement in the foreground dynamic (FD) estimation, the resulting fluctuation in the flow estimation of the static points (FS and BS) is minor.

\subsubsection{Different Loss Combinations}
In this section, as detailed in~\cref{tab:supp_loss}, we present additional ablation studies where we deactivate one of the four losses to analyze the impact of each loss's absence. 
Experiment A2 demonstrates that our model, even without \(\mathcal L_{\text{cham}}\), achieves results comparable to using all loss terms (A1) as suggested in our paper. 
Our dynamic and static losses can replace the general chamfer distance loss to a large extent, but the overall scene-level consideration is still beneficial. 
Our three proposed losses, which are based on dynamic classification and divided into static, dynamic, and object-level aspects, still effectively reduce the EPE 3-way when combined with the Chamfer distance as a foundational constraint.

\begin{table}
\centering
\def\arraystretch{1.3}
\caption{
Ablation study in different loss combinations.
Results are evaluated on the Argoverse 2 \underline{validation set} with 20 training epochs.}
\setlength{\tabcolsep}{1mm}{
\begin{tabular}{ccccc|cccc} 
\toprule
\multirow{2}{*}{Exp. Id} & \multirow{2}{*}{$\mathcal L_{\text{cham}}$} & \multirow{2}{*}{$\mathcal L_{\text{dcham}}$} & \multirow{2}{*}{$\mathcal L_{\text{static}}$} & \multirow{2}{*}{$\mathcal L_{\text{dcls}}$} & \multicolumn{4}{c}{EPE ↓}     \\ 
\hhline{~~~~~----}
& &  &  & & \multicolumn{1}{c}{{\cellcolor[rgb]{0.949,0.949,0.949}}3-way} & \multicolumn{1}{c}{FD} & \multicolumn{1}{c}{FS} & \multicolumn{1}{c}{BS}  \\ 
\hline
A1 & \checkmark & \checkmark & \checkmark & \checkmark & {\cellcolor[rgb]{0.949,0.949,0.949}\textbf{0.0643}} &      0.160    & 0.029             & 0.004 \\
A2 &            & \checkmark & \checkmark & \checkmark & {\cellcolor[rgb]{0.949,0.949,0.949}0.0651}          & 0.162             & 0.030             & 0.003 \\
A3 & \checkmark &            & \checkmark & \checkmark & {\cellcolor[rgb]{0.949,0.949,0.949}0.0717}          & 0.175             & 0.037             & 0.003 \\
A4 & \checkmark & \checkmark &            & \checkmark & {\cellcolor[rgb]{0.949,0.949,0.949}0.0890}          & \textbf{0.150}    & 0.077             & 0.040 \\
A5 & \checkmark & \checkmark & \checkmark &            & {\cellcolor[rgb]{0.949,0.949,0.949}0.0779}          & 0.220             & \textbf{0.012}    & \textbf{0.002}\\

\bottomrule
\end{tabular}
}
\vspace{-0.6em}
\label{tab:supp_loss}
\end{table}

Comparing experiments A3 and A5 with A1 in \cref{tab:supp_loss}, it's evident that omitting \(\mathcal L_{\text{dcham}}\) or \(\mathcal L_{\text{dcls}}\) leads to a decline in dynamic flow estimation accuracy (FD). This highlights the significance of both dynamic and object-level self-supervised losses in assisting networks to understand object motion patterns. Notably, \(\mathcal L_{\text{dcls}}\) (A5) has a more substantial impact than \(\mathcal L_{\text{dcham}}\) (A3). A similar trend is observed for static aspects; comparing A4 with A1 reveals that the absence of \(\mathcal L_{\text{static}}\) results in increased errors in both EPE FS and EPE BS, underscoring its importance in static error reduction.

\subsection{Ablation Study in Difference Model Backbones}
\label{sec:model_backbone}
In this section, we examine the effects of varying the model backbone on performance. We replaced the DeFlow backbone with the FastFlow3D backbone, aligning our model structure with that of ZeroFlow and FastFlow3D. The results, presented in~\cref{tab:model_backbone}, show that even with the same backbone (Ours (FF)), our method still surpasses both ZeroFlow (ZF) and FastFlow3D (FF). This outcome underscores that the strength of our approach lies not in a specific model backbone.

\begin{table}[h]
\centering
\def\arraystretch{1.3}
\caption{
Ablation study in difference model backbones, where FF and DF mean different model backbones from supervised methods FastFlow3D~\cite{fastflow3d} and DeFlow~\cite{zhang2024deflow}, respectively. We \textbf{bold} the best results and \underline{underline} the second best results.
}
\setlength{\tabcolsep}{1.5mm}{
\begin{tabular}{ccccc} 
\toprule
\multirow{2}{*}{BackBone } & \multicolumn{4}{c}{EPE}\\ 
\cline{2-5}
& \multicolumn{1}{c}{{\cellcolor[rgb]{0.949,0.949,0.949}3-way}} & \multicolumn{1}{c}{FD} & \multicolumn{1}{c}{FS} & \multicolumn{1}{c}{BS}  \\ 
\hline
FastFlow3D & {\cellcolor[rgb]{0.949,0.949,0.949}0.081} & 0.222 & 0.020 & 0.002 \\
ZeroFlow (FF) & {\cellcolor[rgb]{0.949,0.949,0.949}0.088} & 0.231  & 0.022     & 0.011  \\
Ours (FF) & {\cellcolor[rgb]{0.949,0.949,0.949}\underline{0.065}} & 0.164    & 0.028     & 0.002  \\
Ours (DF) & {\cellcolor[rgb]{0.949,0.949,0.949}\textbf{0.059}} & 0.147     & 0.026    & 0.004    \\
\bottomrule
\end{tabular}}
\label{tab:model_backbone}
\end{table}

\section{Appendix B - Qualitative Results}
\label{sec:qualitative}
In this section, we present additional qualitative results from the Argoverse 2 and Waymo validation datasets, including some failure cases. 
In each figure, unless otherwise specified, different colors represent different motion directions, and more saturated colors indicate larger flow estimations. 
The qualitative results in the main paper are derived from the scene `b5a7ff7e-d74a-3be6-b95d-3fc0042215f6' in the Argoverse 2 validation set. 
Here, we include two more scenes for further illustration from the Waymo and Argoverse 2 validation set.

\begin{figure}[h]
\centering
\includegraphics[width=\linewidth]{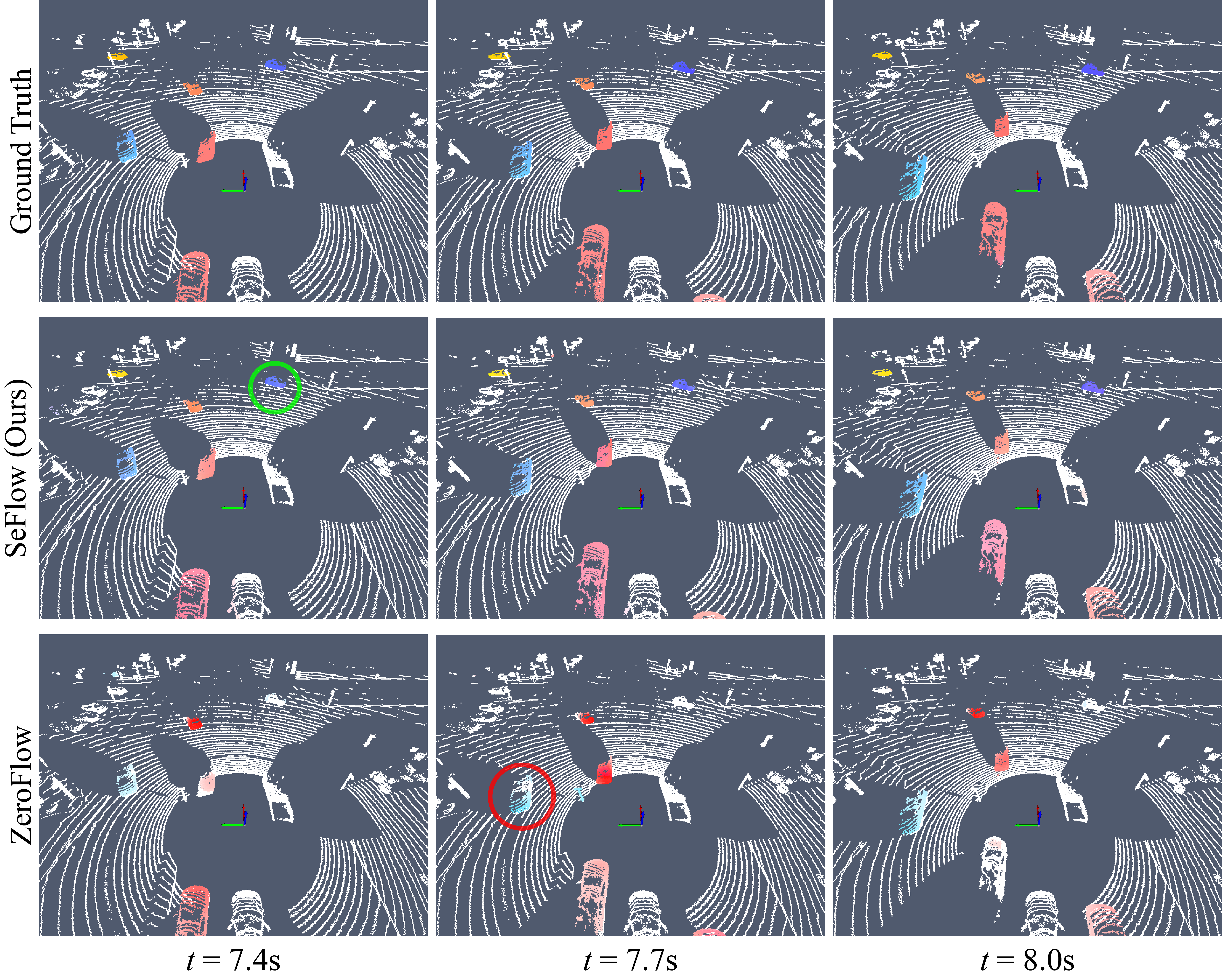}
\caption{
Qualitative results from Waymo validation set (scene id `14081240615915270380\_4399\_000\_4419\_000'). 
The top row displays the ground truth flow, the middle row presents the SeFlow result, and the bottom row showcases the result of another self-supervised method ZeroFlow. 
}
\label{fig:q1_waymo}
\end{figure}

\begin{figure}[h]
\centering
\includegraphics[width=\linewidth]{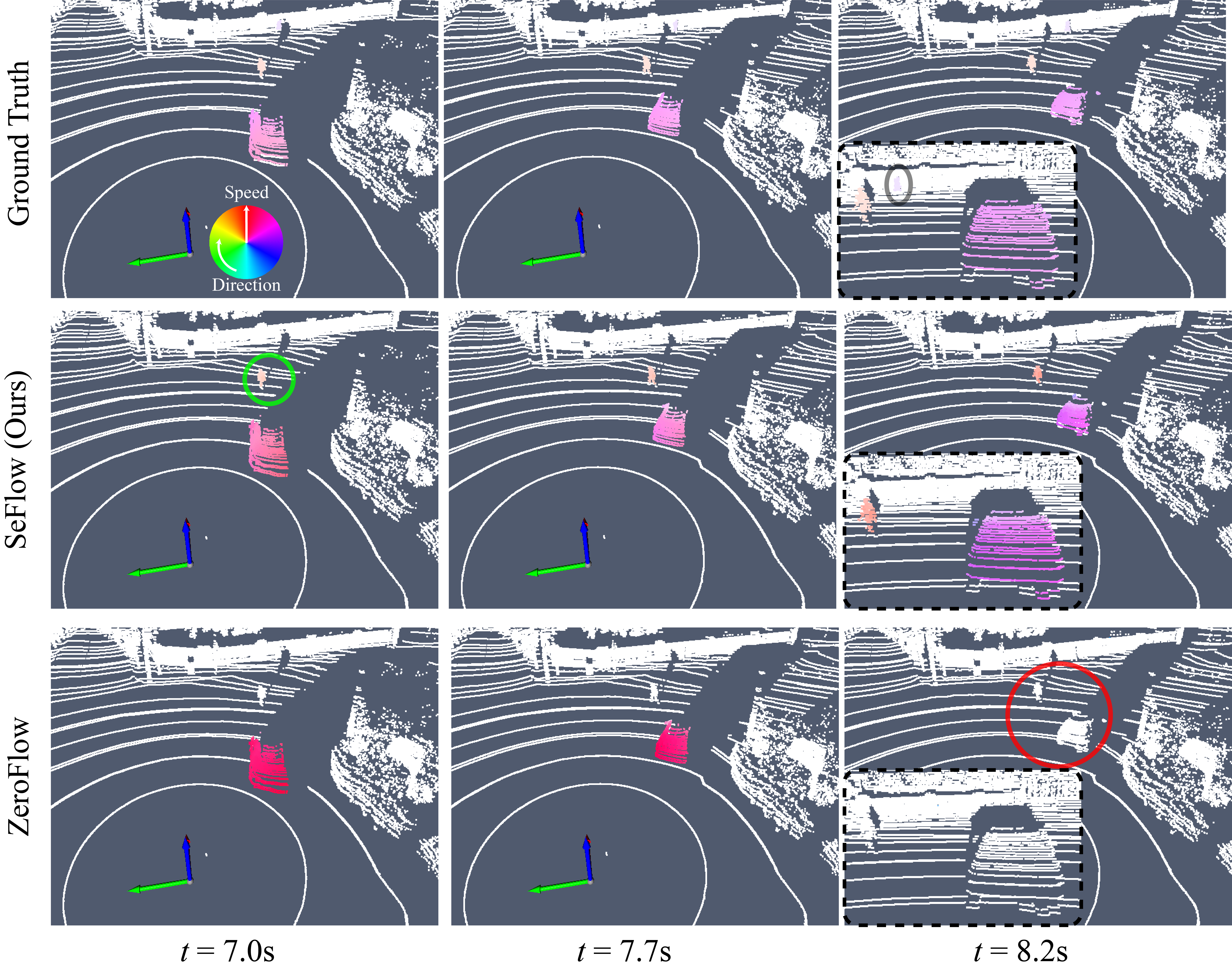}
\caption{Qualitative results from Argoverse 2 validation set (scene id `77574006-881f-3bc8-bbb6-81d79cf02d83'). Different colors represent different motion directions, and more saturated colors indicate larger flow estimations. The top row displays the ground truth flow, the middle row presents the SeFlow result, and the bottom row showcases the result of another self-supervised method ZeroFlow.  The bottom right of the third column is the zoom-in view at the moment. 
}
\label{fig:q2_av2}
\vspace{-2.0em} 
\end{figure}

In terms of flow estimation accuracy, our SeFlow method demonstrates superior performance compared to ZeroFlow in the Waymo dataset, as depicted in~\cref{fig:q1_waymo}. The flows estimated by our method closely align with the ground truth in both direction and magnitude, whereas there are flows from vehicles or parts of vehicles that are ignored in the ZeroFlow results.
In~\cref{fig:q2_av2}, the ZeroFlow results exhibit noticeable issues with no flow estimation in small-scale objects like pedestrians. In contrast, our SeFlow method maintains consistent and accurate flow estimation throughout the scene. 

This additional qualitative analysis further validates the effectiveness of SeFlow in accurately capturing scene dynamics across diverse scenarios.

\subsubsection{Failure Cases}
\begin{figure}[h!]
\centering
\includegraphics[width=\linewidth]{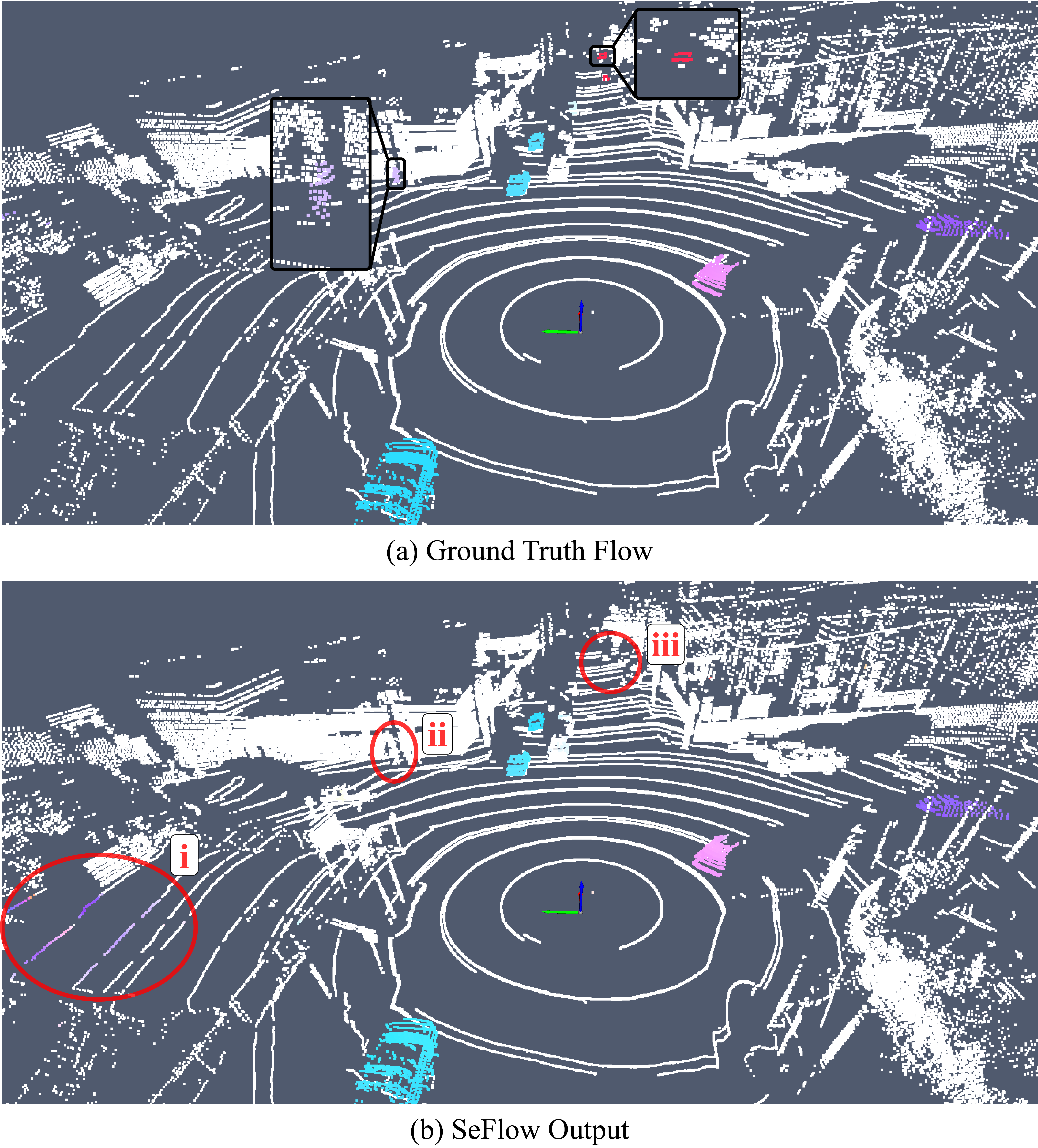}
\caption{Qualitative analysis of failure cases in SeFlow on Argoverse 2 validation set (scene id `22052525-4f85-3fe8-9d7d-000a9fffce36'). (a) displays the ground truth flow where black boxes are the zoom-in views. (b) presents the SeFlow result where the red circle means limitations in our estimation. 
}
\label{fig:q2_av2_limitation}
\end{figure}

As illustrated in \cref{fig:q2_av2_limitation}, our method also has a few deficiencies that need to be improved. One notable issue is the presence of false positive flow estimations, particularly when ground points are not completely removed (\cref{fig:q2_av2_limitation}.b.i). Additionally, predicting the flow of pedestrians near static structures poses a challenge (\cref{fig:q2_av2_limitation}.b.ii). Furthermore, accurately predicting the motion of distant objects proves difficult when relying solely on two consecutive point cloud inputs (\cref{fig:q2_av2_limitation}.b.iii). These limitations highlight specific challenges in scene flow estimation and underscore the need for further refinement and development of our approach.

\end{document}